\pdfoutput=1

\documentclass[11pt]{article}

\usepackage{acl}

\usepackage{times}
\usepackage{latexsym}
\usepackage{subcaption}
\usepackage{booktabs}

\usepackage[T1]{fontenc}

\usepackage[utf8]{inputenc}

\usepackage{microtype}

\usepackage{inconsolata}
\usepackage{amsfonts}
\usepackage{graphicx}
\usepackage{multirow}
\usepackage{caption}
\usepackage[most]{tcolorbox}

\title{\textsc{Community-Cross-Instruct}: Unsupervised Instruction Generation for Aligning Large Language Models to Online Communities}

\author{
Zihao He, Minh Duc Chu$^*$, Rebecca Dorn$^*$, Siyi Guo,  Kristina Lerman\\
USC Information Sciences Institute\\
\texttt{\{zihaoh, mhchu, rdorn, siyiguo\}@usc.edu}, \texttt{lerman@isi.edu}
  }

\begin{document}
\maketitle

\def\thefootnote{\textbf{*}}\footnotetext{Equal Contribution.}\def\thefootnote{\arabic{footnote}}

\begin{abstract}

Social scientists use surveys to probe the opinions and beliefs of populations, but these methods are slow, costly, and prone to biases. Recent advances in large language models (LLMs) enable the creation of computational representations or ``digital twins'' of populations that generate human-like responses mimicking the population's language, styles, and attitudes. We introduce \textsc{Community-Cross-Instruct}, an unsupervised framework for aligning LLMs to online communities to elicit their beliefs. Given a corpus of a community's online discussions, \textsc{Community-Cross-Instruct} automatically generates instruction-output pairs by an advanced LLM to (1) finetune a foundational LLM to faithfully represent that community, and (2) evaluate the alignment of the finetuned model to the community. We demonstrate the method's utility in accurately representing political and diet communities on Reddit. Unlike prior methods requiring human-authored instructions, \textsc{Community-Cross-Instruct} generates instructions in a fully unsupervised manner, enhancing scalability and generalization across domains. This work enables cost-effective and automated surveying of diverse online communities\footnote{Code and data are available at \url{https://github.com/zihaohe123/community-cross-instruct}}.

\end{abstract}

\section{Introduction}
Social scientists use surveys and focus groups to learn the opinions, needs, and concerns of diverse populations. 
However, designing surveys and recruiting participants is a slow and costly process, limiting the utility of these instruments for probing public opinion. Surveys are prone to biases, such as the social desirability bias~\cite{gordon1987social}, where respondents may alter their responses to sensitive questions to appear more socially acceptable \cite{bergen2020everything}, non-response bias~\cite{hill1997non}, where participants fail to answer questions, and self-selection bias due to the choices participants make to participate in the survey~\cite{heckman1990selection}. In addition, social stigmas may taint responses~\cite{goel2010assessing}, especially for hard-to-reach and marginalized groups. 

\begin{figure}[t!]
    \centering
\includegraphics[width=.47\textwidth]{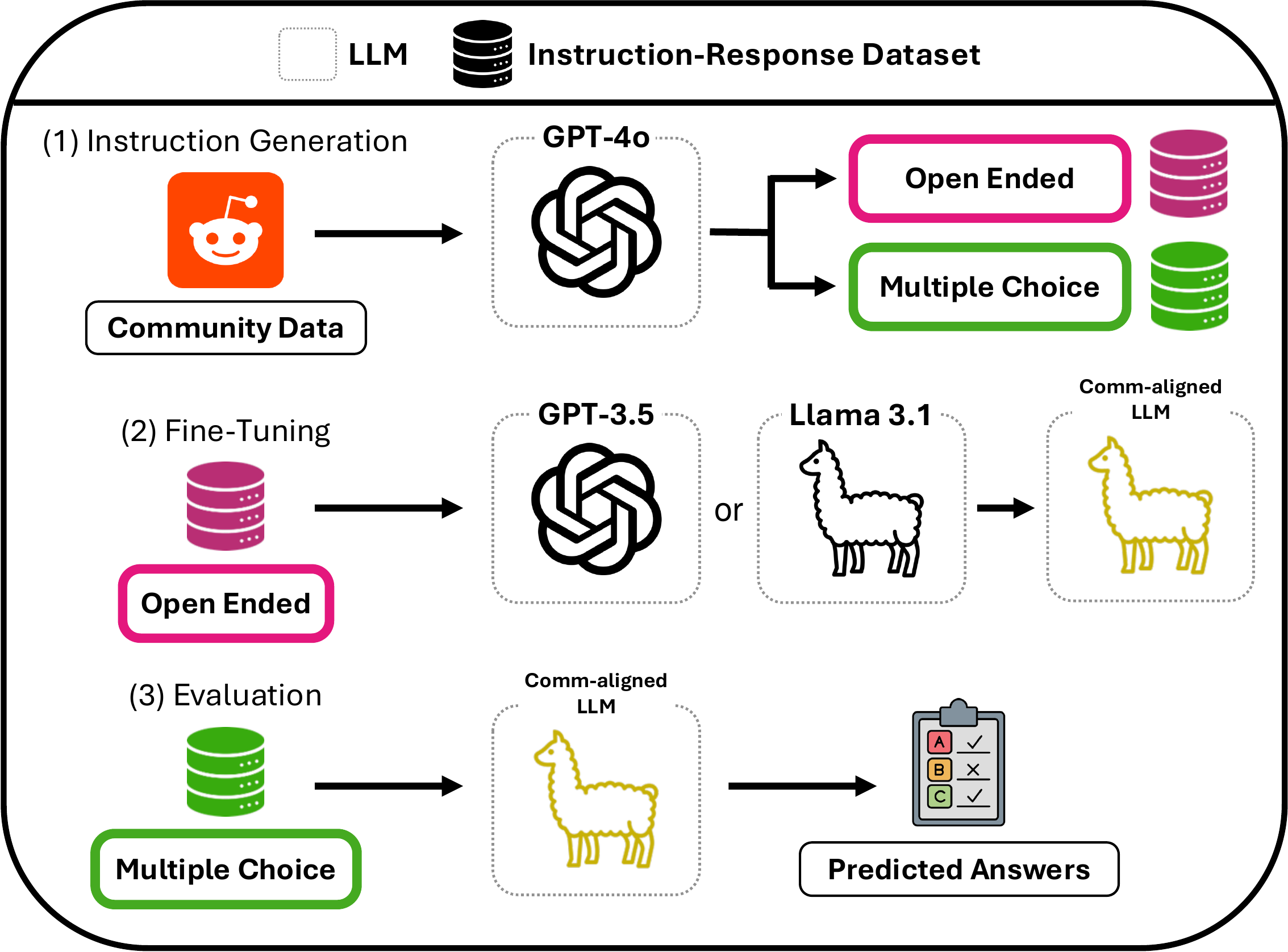}
    \caption{Illustration of \textsc{Community-Cross-Instruct} to align an LLM to a community. (1) Open-ended instructions and multi-choice survey questions are generated by an advanced LLM from the community data. (2) A foundational LLM is aligned to the community through instruction-tuning on the open-ended instructions. (3) The alignment of the finetuned LLM to the community is measured using the generated survey questions.
    }
    \label{fig:overview}
\end{figure}

Recent breakthroughs in generative AI and especially large language models (LLMs) enable new capabilities for creating computational representations of human populations --- their \textit{digital twins}~\cite{el2018digital} --- by ingesting vast textual data they create, for example, in online discussion forums. These LLM-based models generate human-like responses that mimic the language, communication styles, and attitudes of populations they are aligned to, allowing us to probe their worldviews, biases, and sentiments in a cost-effective and automated manner. Previous works have leveraged such LLM-based representations to mine opinions and learn political attitudes of online communities~\cite{jiang2020political,he2024reading}. However, these studies typically finetune models like GPT-2 \cite{radford2019language} directly on raw textual data from the communities, resulting in models that can perform text continuation but are limited in their ability to answer questions in structured formats, such as multiple choice. Moreover, raw community data often contains noise, irrelevant information, or off-topic discussions that can degrade model performance if not properly handled. 



To address these challenges, we aim to align LLMs to communities by finetuning the models with high-quality instructions where the community perspectives are embedded. However, curating high-quality instructional data is a non-trivial process. Existing methods for self-supervised instruction generation rely on a seed set of instructions curated by domain experts \cite{wang2022self, chen2024susceptible}, which is limited by the generalizability to a new domain. 
In this paper, we introduce \textsc{Community-Cross-Instruct}: a fully unsupervised framework for aligning LLMs to online communities through instruction-tuning. It incorporates the readily available community data into the instruction generation pipeline, which no human supervision is needed.
\textsc{Community-Cross-Instruct} uses advanced LLMs to generate instruction-response pairs that better capture community perspectives in more structured and useful formats. These instructions-responses pairs are used to finetune foundational LLMs. 
The finetuned LLMs serve as digital twins of communities, which can be automatically surveyed to elicit their views. We show a high-level overview of our framework in Figure \ref{fig:overview}.

Specifically, given a corpus of comments and submissions in different forums on Reddit, \textsc{Community-Cross-Instruct} uses an advanced LLM (GPT-4o) to automatically curate two instructional datasets: (1) \textsc{CommInst}: a set of 
open-ended instructions  for \textbf{comm}unity-specific \textbf{inst}ruction tuning, and (2) \textsc{CommSurvey}, a set of 
multi-choice survey questions for \textbf{comm}unity-specific \textbf{survey} completion. Each instruction in \textsc{CommInst} and each survey question in \textsc{CommSurvey} are paired with responses from different communities (Figure~\ref{fig:inst_exp}). We finetune foundational LLMs (GPT-3.5 or Llama-3.1) on \textsc{CommInst}, in order to align them to different communities and evaluate the finetuned LLMs on \textsc{CommSurvey} to measure alignment.

Our key contributions can be summarized as
\begin{itemize}
    \item We introduce \textsc{Community-Cross-Inst\-ruct}, a novel unsupervised framework for aligning foundational LLMs to online communities, 
    by finetuning them on the automatically curated set  of open-ended community-specific instruction-response pairs (\textsc{CommInst}). The models' alignment to communities is measured using another set of automatically generated multiple choice questions and answers (\textsc{CommSurvey}).
    \item Using data from Reddit forums, we show that our method improves the fidelity of community representation (alignment) in two domains---\emph{politics} and \emph{diet}---yielding significant alignment improvement over standard persona adaptation methods. 
\end{itemize}

Our work highlights the potential of generative AI to help researchers gain insights from online communities. By leveraging LLMs to create digital twins of these communities, researchers can more accurately and efficiently understand the nuances of public opinion, attitudes, and behaviors. Our framework not only enhances the fidelity of community representation but also paves the way for more effective approaches to studying social phenomena in the digital age.

\paragraph{Using the term \emph{alignment}.} Inspired by \citet{santurkar2023whose}, throughout this paper, we use \emph{alignment} to refer only to the alignment of views and opinions of LLMs and humans.

\section{Problem Definition}
A topical domain (e.g., \emph{politics} or \emph{diet}) includes $n$ communities \{$C_1$, $C_2$, ...., $C_n$\} each with different views and beliefs. Members of each community $C_i$ collectively author text corpus $D_i$ (e.g., discussions on Reddit forums) expressing views and exhibiting behaviors. Our goal is to align an LLM $f$ to each community $C_i$ using its texts $D_i$, such that the aligned LLM $f'_i$ learns the complex mindset of the community and responds to inputs in the community's voice. By administering surveys to the aligned LLMs $\{f'_1, f'_2, ..., f'_n\}$, we obtain responses from different communities, thereby capturing their ideological differences.

\begin{figure}[ht]
    \centering
    \includegraphics[width=0.45\textwidth]{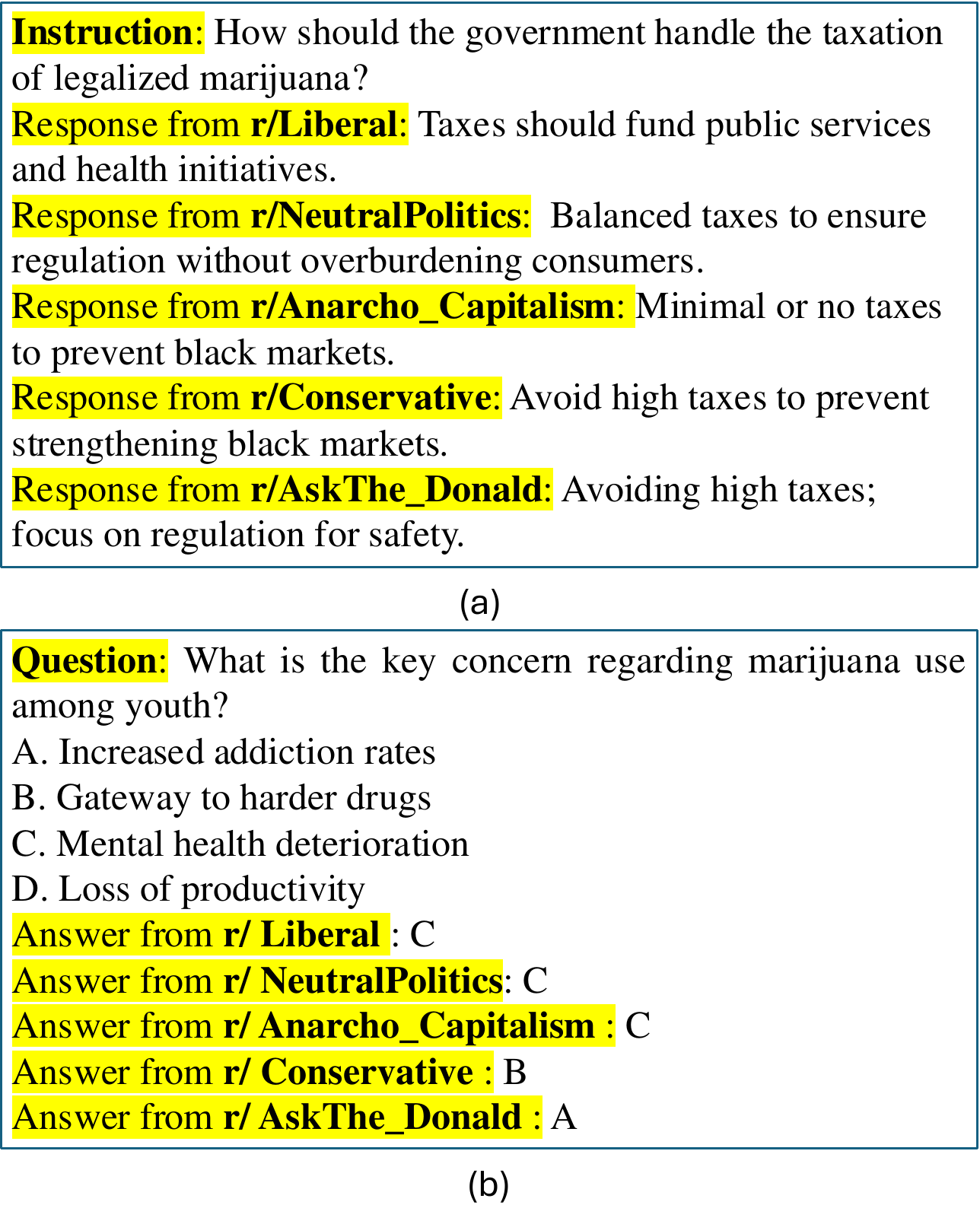}
    \caption{Example of (a) an instruction from \textsc{CommInst} and (b) a survey question from \textsc{CommSurvey} in the politics domain on the topic of marijuana. The open-ended instruction and survey question are paired with answers from different communities.}
    \label{fig:inst_exp}
\end{figure}

\begin{figure*}[ht]
    \centering
    \includegraphics[width=\textwidth]{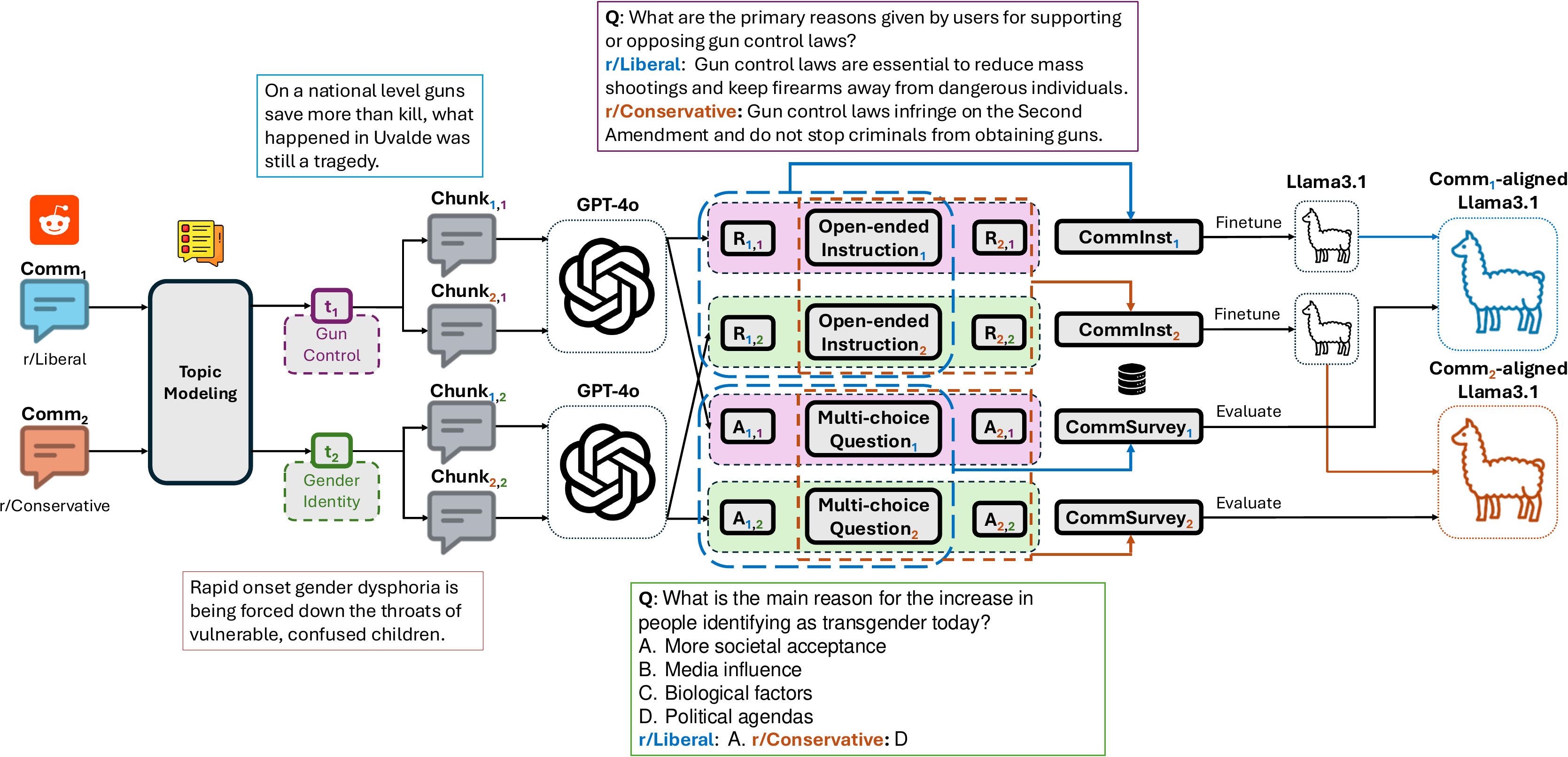}
    \caption{Overview of \textsc{Community-Cross-Instruct}, with an illustrative example of the politics domain. 
    (1) Data is collected for each community within the desired domain. 
    (2) BERTopic clusters the data and identifies prominent topics. A chunk is a set of documents from a community on the same topic. Chunk$_{i,j}$ represents the chunk from community $i$ on topic $j$.
    (3) For each topic, the advanced LLM is prompted with (i) on-topic chunks from each community and (ii) task definition of the instructional data generation (see Appendix \ref{app:prompt_temp_oe}), which leads the LLM to generate (a) open-ended instruction-response pairs and (b) multi-choice question-answer pairs. R$_{i,k}$ represents the response of community $i$ to instruction $k$; A$_{i,k}$ represents the answer of community $i$ to question $k$.
    (4) The open-ended instructions across all topics, along with the corresponding responses of community $i$, are added to \textsc{CommInst}$_i$, which is used to finetune a foundational LLM, to align the LLM to the community.
    (5) The multi-choice questions across all topics, along with the corresponding answers from community $i$, are added to \textsc{CommSurvey}$_i$, which is used to evaluate the finetuned LLM.
    }
    \label{fig:pipeline}
\end{figure*}

To align an LLM $f$ to a community $C$, we finetune it on a set of demonstrations (instruction-response pairs) \cite{wang2023self, ouyang2022training, chen2024susceptible} $I=\{(X_j,Y_j)\}$, where the instructions are open-ended questions probing the community's views on different topics, and the corresponding responses are aligned with each community's ideology. Figure \ref{fig:inst_exp}(a) shows an example demonstration in the politics domain. 
We propose \textsc{Community-Cross-Instruct} (Figure~\ref{fig:pipeline}), a framework to automatically generate community-specific demonstrations $I$ with an advanced LLM $\hat{f}$ (GPT-4o) based on the community's text corpus $D$ and use these demonstrations to instruction-tune a foundational LLM $f$ (GPT-3.5 or Llama-3.1) through a process we call ``\textsc{Cross-Instruct}''.

\section{Online Community Corpora Collection}

Reddit is a vibrant social media platform hosting discussion forums (subreddits) on a wide range of topics \cite{hofmann2022reddit, chen2024isamasred}. 
In this paper, we focus on two domains: \emph{politics} and \emph{diet}, which contain distinctive subreddits, or communities, with complex social dynamics. Political subreddits are valuable expressions of diverse public opinions and viewpoints. In the age of rising polarization, LLMs can help researchers track the complex evolving ideological landscape and effectively elicit public opinions. Meanwhile, the \emph{diet} subreddits feature a wealth of sensitive, health-related conversations, and the latest diet and diet fads, often discussed using inscrutable insider jargon. We can track the direction of these discussions, identifying emerging trends, risks, and potential health misinformation. This insight allows public health officials to promptly address harmful health advice and intervene where necessary. Additionally, it provides an early warning system to detect the spread of dangerous health practices, ensuring that corrective measures can be implemented swiftly to protect community health. 

We identify a set of representative subreddits for each domain based on personal knowledge and by querying ChatGPT. We manually aggregate and filter results, obtaining the following five political online communities: \emph{r/Liberal}, \emph{r/NeutralPolitics},
\emph{r/Anarcho\_Capitalism}, \emph{r/Conservative}, and \emph{r/AskThe\_Donald}; 
For \emph{diet}, we investigate three communities: \emph{r/keto}, \emph{r/WeightLossAdvice}, and \emph{r/EDAnonymous}. More details of these communities are presented in Appendix \ref{app:subreddit}.  

We collect comments and submissions from January 2019 to December 2022 and remove those that were deleted by the author or the moderator through PushShift\footnote{https://pushshift.io}. The statistics of the collected comments and submissions are shown in Appendix \ref{app:subreddit}. We do not differentiate between comments and submissions, and treat each comment and submission as a separate document. In subsequent sections of this paper, we use ``comment'' to refer to both ``comment''
and ``submission''.


\section{Instructional Data Generation}


For each domain, we curate a collection of (1) instructional datasets \textsc{CommInst} for \textbf{comm}unity-specific \textbf{inst}ruction tuning, which is used to finetune LLMs to views of different communities, and (2) survey datasets \textsc{CommSurvey} for evaluating the alignment of the finetuned LLMs' views to the \textbf{comm}unities using \textbf{survey} questions. The pipeline is depicted in Figure \ref{fig:pipeline}.

\subsection{Topic Modeling}

For a domain, denote the combined corpus from all $n$ communities by $\mathbb{D}=D_1 \cup D_2 \cup .... \cup D_n$. We use BERTopic \cite{grootendorst2022bertopic} on $\mathbb{D}$ to identify topics $T$. More details about BERTopic can be found in Appendix \ref{app:bertopic}. After topic modeling, each text $d$ is assigned a topic $t(d)$.  
For each community $C_i$, we split the text corpus $D_i$ into smaller chunks $D_i = \{\hat{D}^i_1, \hat{D}^i_2, ....\}$, where each chunk $\hat{D}^i$ consists of 50 randomly sampled texts belonging to a single topic $t(\hat{D}^i)$. A chunk $\hat{D}^i$ on topic $t(\hat{D}^i)$ is considered an occurrence of the topic.
To investigate different views across the communities, we only keep the topics that are discussed by at least $n-1$ communities. To balance the occurrences of different topics, for each topic in a community, we keep a maximum of 5 chunks. As a result, for \emph{politics} and \emph{diet}, there are 41 and 148 topics respectively. We present the keywords of 10 topics that appear 5 times in all communities in Appendix \ref{app:bertopic}.
\emph{Politics} topics include COVID vaccines, abortion, climate, guns. \emph{diet} topics include body measures, keto diet, and eating disorder recovery.


\subsection{Open-ended Instructions Generation}
\label{sec:inst_ans_gen}
For a specific domain with $n$ communities, we initialize $n$ empty demonstration pools $\{\textsc{CommInst}_i\}_{i=1}^n$. 
For a topic $t_k \in T$ that is discussed by $m$ communities\footnote{As we require a topic to appear at least $n-1$ communities, $m = n$ or $m = n-1$.}, from each community $C_i$, we randomly sample without replacement a text chunk $\hat{D}^i$ on $t_k$.
We include the $m$ sampled text chunks $\{\hat{D}^1, \hat{D}^2, ..., \hat{D}^m\}$ on topic $t_k$ into a prompt and then use the prompt to query the advanced LLM (GPT-4o) to generate open-ended instructions and responses on the topic. The prompting template is shown in Appendix \ref{app:prompt_temp_oe}. In the prompt, we first specify the topic by the topic keywords, and then the comments in the sampled chunks from different communities. Next, we instruct the model to generate an instruction that can be answered based on the texts, and that the instruction should elicit different responses from the communities. In addition, we instruct the model to \textbf{not rely on its pre-knowledge about the community, but solely focus on the given texts}.
The generated instruction $X$ is paired with different responses $\{Y_i\}_{i=1}^m$ from different communities, as shown in Figure \ref{fig:inst_exp}(a). Each demonstration $\{X, Y_i\}$ is then added to the corresponding pool for the $i$-th community.
We iteratively repeat this process until there are fewer than $n-1$ communities that have chunks left on topic $t_k$, then we shift to the next topic $t_{k+1}$. 
After the iterative generation process, each demonstration pool $\textsc{CommInst}_i$ is used to finetune a foundational LLM for community $C_i$. 
Please refer to Appendix \ref{app:stats_comminst} for more details about \textsc{CommInst}.

\subsection{Multi-choice Surveys Generation}
To measure the alignment of the finetuned foundational LLM to the community, we administer a survey of multi-choice questions to the model and evaluate the agreement between the model's responses and the community members' responses. However, manually designing surveys and collecting human responses from online communities is a non-trivial process, which is costly and time-consuming. Instead, we curate another collection of datasets called \textsc{CommSurvey} for evaluating the alignment of the finetuned LLMs' views to the communities. 


We initialize $n$ empty multi-choice question pools $\{\textsc{CommSurvey}_i\}_{i=1}^n$. Then, following the open-ended instruction-response pairs generation process in $\S$\ref{sec:inst_ans_gen}, we iteratively generate multi-choice questions and answers (Figure \ref{fig:inst_exp}(b)) using the same advanced LLM (GPT-4o) and add them to the pools\footnote{In practice we use the advanced LLM to generate open-ended instructions and multi-choice questions at the same time in a single query. For clarity of presentation, we articulate them separately.}. 
The prompting template is shown in Appendix \ref{app:prompt_temp_oe}. Each question pool $\textsc{CommSurvey}_i$ is used to evaluate a finetuned foundational LLM on community $C_i$. Please refer to Appendix \ref{app:stats_comminst} for more details about \textsc{CommSurvey}.

We assume that the answers generated by the advanced LLM are the ``semi-ground truths'', and that they faithfully represent the views of the corresponding communities. 
As an empirical evaluation, we compute the pairwise agreement between different communities using their responses to questions in $\textsc{CommSurvey}$. The results are presented in Figure \ref{fig:agreement_mat} in Appendix \ref{app:comm_agreement}.
We observe significant polarization between the left-leaning and right-leaning communities, and strong agreement between the two right-leaning communities (\emph{r/Conservative} and \emph{r/AskThe\_Donald}). However, in the \emph{diet} domain where the communities are more practical, there is less polarization. 
For more rigorous evaluation, we verify the ``semi-ground truths'' by human annotation, as detailed in Appendix \ref{app:human_anno}.

\section{Experiments}
\label{sec:exp}
We align foundational LLMs to Reddit communities by finetuning them on relevant demonstrations from \textsc{CommInst}. After finetuning, we evaluate the models using surveys from \textsc{CommSurvey}, comparing their responses to ``semi-ground truths'' to assess alignment.

\subsection{Experimental Setup}

\paragraph{Data Generation.} We use OpenAI's GPT-4o Batch API to generate the datasets. For each query, three open-ended instructions and two multi-choice questions are created, as shown in Appendix \ref{app:prompt_temp_oe}. The API costs approximately \$2 for the \emph{politics} domain and \$5 for the \emph{diet} domain.

\paragraph{Data Splitting.}
Among all the queries used to prompt the advanced LLM to generate data, we randomly select 85\% of them and use the instructions generated from these queries as the training instructions denoted as \textsc{CommInst-train}. For the rest 15\% queries, we use the corresponding multi-choice questions as test questions, denoted as \textsc{CommSurvey-test}. In addition to randomly splitting the queries, we perform a topic-wise split, and make sure that the 85\% training instructions, \textsc{CommInst-train-topic}, and 15\% test questions, \textsc{CommSurvey-test-topic}, do not cover the same topics.

\paragraph{Finetuning.}
We focus on two strong foundational LLMs -- Llama-3.1-8B-Instruct \cite{dubey2024llama} and GPT-3.5-Turbo \cite{ouyang2022training}.
For each community $C_i$, we finetune the LLM on \textsc{CommInst-train$_i$} ($\S$\ref{sec:main_res}) and \textsc{CommInst-train-topic$_i$} ($\S$\ref{sec:oot_res}). The input and the output are the instruction and response verbatim.
Llama-3.1 is finetuned with \textsc{LlamaFactory} \cite{zheng2024llamafactory}, using both full and LoRA finetuning \cite{hu2022lora}, with batch size 16 for 3 epochs. We report the results of the model (full or LoRA) that achieves better loss on the validation set (5\% of the training data). Full finetuning takes around 3 minutes on 4 NVIDIA H100 GPUs, and LoRA finetuning takes around 30 seconds on 1 GPU. 
For GPT-3.5, we use the OpenAI API, which completes finetuning in around 10 minutes for 1 epoch, where the batch size is automatically determined.

\paragraph{Measuring Alignment.}
We administer survey questions from \textsc{CommSurvey-test} ($\S$\ref{sec:main_res}) and \textsc{CommSurvey-test-topic} ($\S$\ref{sec:oot_res}) to the finetuned foundational LLMs. 
Each prompt includes the question and options verbatim, followed by the instruction, ``Select only one answer by stating either A, B, C, or D. Do not provide any additional explanation or rationale for your choice,'' to facilitate easier matching of the model’s responses to the options.
We set the temperature to 0.8, generating 20 responses per question, and take the majority vote. Accuracy is calculated by comparing the model's answers to the ``semi-ground truths''.


\begin{table*}[ht]
\centering
\small
\def\arraystretch{1.2}
\addtolength{\tabcolsep}{-3.0pt}
\begin{tabular}{lcccccccc}
\hline
\multirow{3}{*}{\textbf{Subreddit}} & \multicolumn{4}{c}{\textbf{Llama-3.1-8B}}                                                                                                                               & \multicolumn{4}{c}{\textbf{GPT-3.5-Turbo}}                                                                                                                              \\ \cline{2-9} 
                                    & \multicolumn{2}{c}{\textbf{Without Steering}}                                 & \multicolumn{2}{c}{\textbf{With Steering}}                                              & \multicolumn{2}{c}{\textbf{Without Steering}}                                 & \multicolumn{2}{c}{\textbf{With Steering}}                                              \\
                                    & \multicolumn{1}{l}{\textbf{Context}} & \multicolumn{1}{l}{\textbf{CrossInst}} & \multicolumn{1}{l}{\textbf{Steering}} & \multicolumn{1}{l}{\textbf{Steering+CrossInst}} & \multicolumn{1}{l}{\textbf{Context}} & \multicolumn{1}{l}{\textbf{CrossInst}} & \multicolumn{1}{l}{\textbf{Steering}} & \multicolumn{1}{l}{\textbf{Steering+CrossInst}} \\ \hline
r/Liberal                           & 8.3                                  & \multicolumn{1}{c|}{\textbf{54.2}}     & 55.8                                  & \multicolumn{1}{c|}{\textbf{58.3}}              & 41.7                                 & \multicolumn{1}{c|}{\textbf{62.5}}     & 45.8                                  & \textbf{62.5}                                   \\
r/NeutralPol                        & 3.8                                  & \multicolumn{1}{c|}{\textbf{50.0}}     & 55.0                                  & \multicolumn{1}{c|}{\textbf{63.3}}              & \textbf{55.0}                        & \multicolumn{1}{c|}{\textbf{55.0}}     & 40.0                                  & \textbf{50.0}                                   \\
r/Anar\_Cap                         & 54.1                                 & \multicolumn{1}{c|}{\textbf{76.7}}     & \textbf{70.0}                         & \multicolumn{1}{c|}{66.7}                       & 50.0                                 & \multicolumn{1}{c|}{\textbf{66.7}}     & 66.7                                  & \textbf{73.3}                                   \\
r/Conservative                      & 18.4                                 & \multicolumn{1}{c|}{\textbf{63.3}}     & \textbf{60.5}                         & \multicolumn{1}{c|}{56.7}                       & 50.0                                 & \multicolumn{1}{c|}{\textbf{53.3}}     & \textbf{53.3}                         & \textbf{53.3}                                   \\
r/AT\_Donald                        & 23.7                                 & \multicolumn{1}{c|}{\textbf{56.7}}     & 66.7                                  & \multicolumn{1}{c|}{\textbf{70.0}}              & 30.0                                 & \multicolumn{1}{c|}{\textbf{56.6}}     & 50.0                                  & \textbf{63.3}                                   \\
\textbf{avg. politics}              & 21.7                                 & \multicolumn{1}{c|}{\textbf{60.2}}     & 61.6                                  & \multicolumn{1}{c|}{\textbf{63.0}}              & 45.3                                 & \multicolumn{1}{c|}{\textbf{58.8}}     & 51.2                                  & \textbf{60.5}                                   \\ \hline
r/keto                              & 26.0                                 & \multicolumn{1}{c|}{\textbf{75.0}}     & 67.0                                  & \multicolumn{1}{c|}{\textbf{72.0}}              & 56.0                                 & \multicolumn{1}{c|}{\textbf{68.0}}     & \textbf{66.0}                         & \textbf{66.0}                                   \\
r/WLAdvice                          & 27.7                                 & \multicolumn{1}{c|}{\textbf{60.6}}     & \textbf{65.2}                         & \multicolumn{1}{c|}{61.7}                       & \textbf{59.6}                        & \multicolumn{1}{c|}{58.5}              & \textbf{66.1}                         & 64.8                                            \\
r/EDAnony                           & 29.1                                 & \multicolumn{1}{c|}{\textbf{72.1}}     & 65.1                                  & \multicolumn{1}{c|}{\textbf{69.8}}              & 61.6                                 & \multicolumn{1}{c|}{\textbf{72.1}}     & 62.7                                  & \textbf{66.3}                                   \\
\textbf{avg. diet}                  & 27.6                                 & \multicolumn{1}{c|}{\textbf{69.2}}     & 65.8                                  & \multicolumn{1}{c|}{\textbf{67.8}}              & 59.1                                 & \multicolumn{1}{c|}{\textbf{66.2}}     & 64.9                                  & \textbf{65.7}                                   \\ \hline
\end{tabular}
\addtolength{\tabcolsep}{3.0pt}
\caption{Evaluation results of Llama-3.1-8B and GPT-3.5-Turbo on \textsc{CommSurvey-test}. The community-aligned LLMs (\textsc{CrossInst} and \textsc{Steering}+\textsc{CrossInst}) are finetuned on \textsc{CommInst-train}, so there is potential topic overlap between training and evaluation. For each model family, the results are divided into two groups, one without and one with steering. The best results in each group are highlighted in bold.
}
\label{tab:main_res}
\end{table*}

\subsection{Baselines}

\paragraph{LLM+Context.}
For each community $C_i$, we provide the vanilla (unfinetuned) LLM with the context about the community by appending 300 most relevant comments from it. For example, when prompting the LLM to answer a question in \textsc{CommSurvey-test$_i$}, we retrieve the most relevant 300 comments to the question, from the text chunks that are used to generate \textsc{CommSurvey-train$_i$}, by calculating the embedding similarity with \emph{sentence-transformers}. The prompt is augmented with the instruction ``\texttt{According to the following statements, learn the mindset and select only one most relevant answer by stating either A, B, C, or D. Do not provide any additional explanation or rationale for your choice. [comments]}''. This baseline is inspired by the idea of in-context learning, where the LLM learns the community's mindset within the context provided in the prompt.

Providing the LLM with context becomes less efficient when the model is deployed to answer a large number of survey questions, as each question requires processing an extremely long input sequence due to the added comments. This significantly increases computational costs, memory usage, and processing time, making it inefficient for large-scale or real-time applications. In contrast, our framework, which finetunes the model on community-specific data, is a one-time effort. Once the model is finetuned, it can efficiently answer survey questions without requiring additional long contextual inputs, making it far more scalable and resource-friendly in practical use cases.

\paragraph{LLM+Steering.} When prompting the vanilla LLM to answer survey questions, we steer it to mimic the community, by specifying in the prompt that ``\texttt{Select only one answer \textbf{that best aligns with the opinions of members from subreddit r/[subreddit]}}''. Steering the vanilla LLMs can nudge them to respond to the community. 
For fair comparison to this baseline, we also steer the finetuned LLM via \textsc{Community-Cross-Instruct}, where the LLM is aligned to the community both in the finetuning and steering process.

It is worth noting that \textbf{steering only applies to predefined communities that are developed via a manually specified tag}, such as subreddits, where we can easily reference community names in the prompt.
While this paper focuses on forum-based communities, \textsc{Community-Cross-Instruct} generalizes to other online communities, including organically formed communities in the retweet network \cite{chu2024characterizing} or the news co-sharing network \cite{mosleh2021cognitive, he2024reading}, as long as their relevant text data is readily available.
Although it is not always feasible to concisely summarize organically formed communities in text, \textsc{Community-Cross-Instruct} allows for LLM alignment to such communities without requiring explicit textual descriptions.

\begin{table*}[ht]
\centering
\small
\def\arraystretch{1.2}
\addtolength{\tabcolsep}{-3.0pt}
\begin{tabular}{lcccccccc}
\hline
\multirow{3}{*}{\textbf{Subreddit}} & \multicolumn{4}{c}{\textbf{Llama-3.1-8B}}                                                                                                                               & \multicolumn{4}{c}{\textbf{GPT-3.5-Turbo}}                                                                                                                              \\ \cline{2-9} 
                                    & \multicolumn{2}{c}{\textbf{Without Steering}}                                 & \multicolumn{2}{c}{\textbf{With Steering}}                                              & \multicolumn{2}{c}{\textbf{Without Steering}}                                 & \multicolumn{2}{c}{\textbf{With Steering}}                                              \\
                                    & \multicolumn{1}{l}{\textbf{Context}} & \multicolumn{1}{l}{\textbf{CrossInst}} & \multicolumn{1}{l}{\textbf{Steering}} & \multicolumn{1}{l}{\textbf{Steering+CrossInst}} & \multicolumn{1}{l}{\textbf{Context}} & \multicolumn{1}{l}{\textbf{CrossInst}} & \multicolumn{1}{l}{\textbf{Steering}} & \multicolumn{1}{l}{\textbf{Steering+CrossInst}} \\ \hline
r/Liberal                           & 11.1                                 & \multicolumn{1}{c|}{\textbf{50.0}}     & 58.3                                  & \multicolumn{1}{c|}{\textbf{61.1}}              & 41.7                                 & \multicolumn{1}{c|}{\textbf{61.1}}     & 63.9                                  & \textbf{65.1}                                   \\
r/NeutralPol                        & 20.8                                 & \multicolumn{1}{c|}{\textbf{60.0}}     & 54.2                                  & \multicolumn{1}{c|}{\textbf{65.0}}              & 52.5                                 & \multicolumn{1}{c|}{\textbf{65.0}}     & \textbf{67.5}                         & \textbf{67.5}                                   \\
r/Anar\_Cap                         & 23.9                                 & \multicolumn{1}{c|}{\textbf{56.5}}     & 52.2                                  & \multicolumn{1}{c|}{\textbf{56.5}}              & 54.3                                 & \multicolumn{1}{c|}{\textbf{58.7}}     & 58.7                                  & \textbf{71.7}                                   \\
r/Conservative                      & 20.0                                 & \multicolumn{1}{c|}{\textbf{54.3}}     & 52.5                                  & \multicolumn{1}{c|}{\textbf{65.2}}              & 50.0                                 & \multicolumn{1}{c|}{\textbf{67.4}}     & \textbf{69.6}                         & 65.2                                            \\
r/AT\_Donald                        & 25.0                                 & \multicolumn{1}{c|}{\textbf{56.5}}     & \textbf{57.5}                         & \multicolumn{1}{c|}{56.5}                       & 52.2                                 & \multicolumn{1}{c|}{\textbf{60.9}}     & 61.2                         & \textbf{63.0}                                            \\
\textbf{avg. politics}              & 20.2                                 & \multicolumn{1}{c|}{\textbf{55.5}}     & 54.9                                  & \multicolumn{1}{c|}{\textbf{60.9}}              & 50.1                                 & \multicolumn{1}{c|}{\textbf{62.6}}     & 64.2                                  & \textbf{66.5}                                   \\ \hline
r/keto                              & 29.7                                 & \multicolumn{1}{c|}{\textbf{64.4}}     & \textbf{65.2}                         & \multicolumn{1}{c|}{63.6}                       & 59.3                                 & \multicolumn{1}{c|}{\textbf{67.8}}     & 61.0                                  & \textbf{63.5}                                   \\
r/WLAdvice                          & 26.9                                 & \multicolumn{1}{c|}{\textbf{63.5}}     & 61.5                                  & \multicolumn{1}{c|}{\textbf{62.5}}              & 58.7                                 & \multicolumn{1}{c|}{\textbf{62.5}}     & \textbf{63.5}                         & 62.5                                            \\
r/EDAnony                           & 25.4                                 & \multicolumn{1}{c|}{\textbf{55.9}}     & 50.0                                  & \multicolumn{1}{c|}{\textbf{54.9}}              & 57.8                                 & \multicolumn{1}{c|}{\textbf{60.8}}     & 58.8                                  & \textbf{61.8}                                   \\
\textbf{avg. diet}                  & 27.3                                 & \multicolumn{1}{c|}{\textbf{61.3}}     & 58.9                                  & \multicolumn{1}{c|}{\textbf{60.3}}              & 58.6                                 & \multicolumn{1}{c|}{\textbf{63.7}}     & 61.1                                  & \textbf{62.6}                                   \\ \hline
\end{tabular}
\addtolength{\tabcolsep}{3.0pt}
\caption{Evaluation results of Llama-3.1-8B and GPT-3.5-Turbo on \textsc{CommSurvey-test-topic}. The community-aligned LLMs (\textsc{CrossInst} and \textsc{Steering}+\textsc{CrossInst}) are finetuned on \textsc{CommInst-train-topic}, so there is \textbf{no topic overlap} between training and evaluation. For each model family, the results are divided into two groups, one without steering and one with steering. The best results in each group are highlighted in bold.}
\label{tab:oot-res}
\end{table*}

\subsection{Main Results}
\label{sec:main_res}

We compare the finetuned LLMs' generated answers to multi-choice questions to the ``semi-ground truths'' and report the accuracy in Figure \ref{tab:main_res} as a measure of alignment with the corresponding community. The LLMs are finetuned on \textsc{CommInst-train} and evaluated on \textsc{CommInst-test}. 

\paragraph{Politics.} In the \emph{politics} domain, \textsc{CrossInst} consistently outperforms \textsc{Context} across both LLMs, demonstrating the strength of finetuning on community-specific instructional data. By aligning LLMs to explicit community instructions, \textsc{CrossInst} enables the models to better capture the nuances of political discourse, where ideological divides and subtle differences in language are critical. This makes the model more adept at accurately reflecting community values without being overwhelmed by noisy or redundant information, which is a common issue with \textsc{Context}. We observe that the a large portion of generated answers from Llama-3.1 using \textsc{Context} are texts irrelevant to the survey questions, indicating that it struggles in dealing with a lengthy prompt with 300 examples.

An important aspect to consider is that LLMs come with pre-existing knowledge about various communities, which is learned during pretraining on large-scale internet data. During \textsc{Steering}, the model attempts to retrieve and apply this pre-existing knowledge to align its responses with the given community. However, this knowledge may not always be fully accurate or consistent with the actual values of the community being evaluated. For example, \textsc{Steering} might prompt the model to draw on generalizations or stereotypes that are present in its pretraining data, which might not reflect the current or specific views of the subreddit in question. This inconsistency can explain why \textsc{Steering} alone sometimes leads to suboptimal performance.

When combining \textsc{Steering} with \textsc{CrossInst}, a potential conflict arises between the knowledge gained from finetuning and the pre-existing community knowledge the model tries to retrieve through steering. The model is essentially being pulled in two directions: one based on the explicit, finetuned instructions from \textsc{CrossInst} and the other based on its internalized, sometimes outdated, pretraining knowledge. This conflict can result in \textsc{Steering}+\textsc{CrossInst} underperforming compared to \textsc{Steering} alone, as observed in certain subreddits, such as \emph{r/Anarcho\_Capitalism} on Llama-3.1. In these cases, the inconsistency between the steering prompts and the finetuned knowledge creates confusion for the model, leading to lower performance.

\paragraph{Diet.} In the \emph{diet} domain, we observe a similar pattern where \textsc{CrossInst} significantly outperforms the \textsc{Context} baseline. Communities like \emph{r/keto} and \emph{r/EDAnonymous}, which are focused on specific health and lifestyle goals, benefit greatly from instructional finetuning. These communities are characterized by practical and focused discussions, and \textsc{CrossInst} allows the model to adapt to the specific language and norms within these subreddits, ensuring a more accurate reflection of their values and objectives. 

Interestingly, in \emph{r/WeightLossAdvice}, \textsc{Steering} alone performs the best across the four methods within each LLM. One possible reason for this is that in certain communities, particularly those that are well-represented in pretraining data, the model’s pre-existing knowledge might be more aligned with the actual community values. In such communities, where strong ideological markers may already be embedded in the model's pretraining data, steering allows the model to retrieve this relevant information effectively, leading to stronger performance. In such cases, \textsc{Steering} helps amplify the model’s pre-learned alignment with the community, which can be sufficient for capturing the community’s voice without the need for additional finetuning.

\subsection{Out-of-Topic Generalizability}
\label{sec:oot_res}
Our final goal is to create LLMs aligned to different communities, which can be used to answer any survey question from the perspective of the community. In the real world, the surveys may contain questions covering topics that do not appear in the finetuning data \textsc{CommInst}. In this study, we finetune the LLMs on \textsc{CommInst-train-topic}, and evaluate them on \textsc{CommSurvey-test-topic}, to make sure that there is no overlap in topic coverage. 

The results are shown in Table \ref{tab:oot-res}.
\textsc{CrossInst} models continue to outperform the \textsc{Context} baselines, demonstrating their strong generalization capability to new topics, suggesting that \textsc{CrossInst} maintains its robustness even when exposed to entirely new topics. 

We observe that \textsc{Steering} alone occasionally performs the best (e.g., \emph{r/AskThe\_Donald} on both LLMs, and \emph{r/WeightLossAdvice} on GPT-3.5). We argue that this might be because ideologically-polarized or practical communities are consistent in their mindsets over time, so the LLM's pre-knowledge about them would well predict their ideologies in the future.




\section{Related Works}
\subsection{Self-Improved Instruction Tuning.}
A survey of instruction-tuning approaches \cite{zhang2023instruction} outlines several methods for models to autonomously self-improve their instruction set. These include generating instructions for pre-existing texts \cite{li2023self}, eliciting interaction between different model iterations \cite{chen2024self}, and bootstrapping from an existing instruction set to generate new ones \cite{wang2022self, chen2024susceptible}. Additionally, a human-in-the-loop framework builds upon self-generalization by iterating between human and machine-generated instructions \cite{guo2024human}. Despite these advances, these strategies require manual input. In this work, we build on \textit{self-instruct}, a method to enhance instruction tuning by eliciting synthetic model instruction generations. We alter the original framework by removing the required set of manually written seed instructions \cite{wang2022self}. This work significantly contributes to the field by offering a scalable and fully autonomous solution to instruction tuning, paving the way for more adaptive and intelligent models.

\subsection{Aligning LLMs to Subgroups}
Existing work has aligned LLMs to different human subgroups to discover their mindset and opinions \cite{dorn2023non, dorn2024harmful}. 
Subpopulation representative models (SRMs) \cite{simmons2023large} can be used to emulate some characteristics of a particular subpopulation, particularly as LLMs can provide fine-grained, demographically-correlated outputs \cite{Argyle_Busby_Fulda_Gubler_Rytting_Wingate_2023}. 
\citet{argyle2023out} find that exposing GPT-3 to thousands of socio-demographic backstories leads the model to obtain a  complex understandings of sociological concepts.

To learn about partisan communities, \citet{jiang2022communitylm} propose \textsc{CommunityLM} by finetuning GPT-2 on tweets authored by prominent community members, and prompt the finetuned model to generate opinions about various political entities. \citet{he2024reading} extends \textsc{CommunityLM} to organically-formed online communities with more fine-grained ideologies. However, these finetuned GPT-2 models can only be used for text continuation tasks and cannot answer survey questions. In our paper, we finetune the foundational instruction-tuned LLMs to serve as their digital twins. The resulting LLMs retain the instruction-following capabilities and are able to complete various tasks as specified in the instructions, including survey completion.

\subsection{Evaluating LLMs' Subgroup Alignment}
Social scientists use surveys to systematically collect data from populations to characterize their beliefs, attitudes, opinions, and behaviors \cite{hill1997non, choi2005peer, gordon1987social, goel2010assessing}. 
LLM developers and practitioners focus on measuring LLM’s alignment with different human subgroups \cite{santurkar2023whose, durmus2023towards, he-etal-2024-whose} using use real-world survey responses as ground truth for comparing values. However, they often ignore under-represented groups, as it is difficult to administer surveys to people from those groups. To administer surveys, researchers sample individuals from a target population and ask them to manually respond to survey questions. This can be costly and slow, which limits the utility and timeliness of surveys. 
Our method addresses this challenge by using social media data that is easily accessible, even for those hard-to-reach subgroups. By automatically creating multi-choice survey questions and answers \textsc{CommSurvey} for these groups, our method enables a more comprehensive and cost-effective evaluation of LLMs’ alignment with diverse groups.


\section{Discussions}
We aim to align LLMs to online communities through instruction-tuning so that the aligned LLMs can be flexibly surveyed to probe the communities' views on different topics. To prepare data for instruction tuning, we propose \textsc{Community-Cross-Instruct}, and the automatic creation of the open-ended instruction-response pairs (\textsc{CommInst}) is the core contribution. However, in order to demonstrate that the finetuned LLMs are indeed aligned to communities and can be treated as their digital twins, we propose to automatically generate survey questions and answers (\textsc{CommSurvey}) to efficiently evaluate LLM alignment. This is a secondary contribution of this paper, and we do not claim that  \textsc{CommSurvey} is the best way to measure alignment. Therefore, our main goal is to propose a method to prepare LLMs as a proxy for surveying different online communities, rather than automatically creating such surveys.



\section{Conclusion}
In this paper, we introduced and explored \textsc{Community-Cross-Instruct}, a fully automated framework to represent online communities and elicit their views. By automating the process of surveying diverse online communities, \textsc{Community-Cross-Instruct} offers a cost-effective and efficient alternative to traditional survey methods used by social scientists. Through finetuning LLMs based on community-specific instructions and answers, \textsc{Community-Cross-Instruct} enables the generation of accurate responses that align with the beliefs and perspectives of different online communities. This innovative framework has demonstrated its effectiveness in representing various communities, including political and diet forums on Reddit. \textsc{Community-Cross-Instruct} opens up new possibilities for researchers to gain insights into online communities' diverse views and opinions, paving the way for deeper understanding and analysis of digital societies.

Moving forward, we will apply the framework to model organic online communities, such as those in the retweet network. In addition, we are interested in aligning LLMs to communities through reinforcement learning from human feedback (RLHF).

\section*{Limitations}
\paragraph{Ignoring Thread Structure in Reddit Data.}
We treat each Reddit comment and submission as independent data points, without taking into account the thread structure in which they are embedded. Reddit discussions are inherently hierarchical, with comments often building upon or responding to previous comments and submissions. By ignoring this structure, we may lose important contextual information that could help capture the nuances of community interactions and opinions. Future work could explore leveraging the thread structure to better model the flow of conversations, capturing how community members engage with each other’s ideas and how opinions evolve within discussions.

\paragraph{Identical LLM for Generating both Datasets.} 
We use GPT-4o for generating both \textsc{CommInst} for finetuning LLMs and \textsc{CommSurvey} for evaluating LLM alignment, as we couldn't find a different advanced LLM that was able to generate high-quality data as GPT-4o. This reliance on a single LLM introduces a limitation, as it may lead to a bias where the evaluation dataset (CommSurvey) is inherently aligned with the model used for instruction generation (CommInst), potentially overestimating the alignment accuracy of the finetuned models. To address this limitation, we plan to incorporate a diverse set of advanced LLMs as they become available in future iterations of our framework. Additionally, we will include human-in-the-loop validation to ensure the generated datasets maintain high quality and representativeness, mitigating any biases introduced by the single LLM dependency. Furthermore, we will explore cross-validation techniques and third-party evaluations to benchmark the performance of our finetuned models, ensuring the robustness and generalizability of our results beyond the influence of a single LLM.

\paragraph{Group Approximation.}
To approximate group-level behavior, community members represented in the minority might be inherently excluded. 
Further, measuring group alignment using a single-answer multiple choice questionnaire does not account for a more complex distribution of the various opinions within the community. We hope to build on this work by experimenting with survey designs that account for more diversity in communities.

\paragraph{Hallucination Potential.}
Adapting models to communities poses a risk of language models hallucinating or providing misinformation in their community representation.
We hope to build upon this work with in-depth experiments on model hallucination and misrepresentation in the subgroup representation task.

\paragraph{Prompt Perturbations.}
LLM responses are sensitive to slight changes in prompts \cite{salinas2024butterfly}. In this work, we work primarily with one prompt for instruction generation. We would be interested to see how the model's generated instructions shift with different prompting schema.

\section*{Ethics Statement}
\paragraph{Finetuning LLMs towards Bias.}
Aligning LLMs with specific communities may result in models that appear more biased, as they are finetuned to the distinctive views and perspectives of those communities. However, this process is done solely for the purpose of accurately reflecting the values and attitudes prevalent within each community, rather than intentionally making the models more biased. Our goal is not to reinforce or amplify harmful biases, but rather to provide a computational tool that can represent the views of a given community for research purposes. To mitigate potential misuse, these models should be used in controlled environments and in contexts where understanding specific community perspectives is essential for social, cultural, or political research.

\paragraph{User Privacy and Consent.} 
Users might not be informed of how their data (reflected in their posts) are used to produce digital twins that mirror their voices and the purpose of the survey constructed based on their data. Furthermore, automatically surveying communities can reveal unconsented insights of certain individuals or groups of people online. To address these ethical concerns, we implement several measures. Firstly, we ensure that all data used for creating digital twins is anonymized, stripping any personally identifiable information to protect user privacy. Secondly, we seek to aggregate data in a manner that focuses on community-level insights rather than individual-level analysis, thereby reducing the risk of unconsented personal exposure. Thirdly, we will obtain explicit consent from users where possible, clearly communicating the purpose and scope of our research. Lastly, we will adhere to ethical guidelines and institutional review board (IRB) requirements to ensure that our methods respect the privacy and rights of the individuals and communities involved.

\section*{Acknowledgements}
We are grateful to Patrick Gerard, Ashwin Rao, and DJ Berry for helping review responses to surveys. This project was funded in part by the Defense Advanced Research Projects Agency (DARPA)  under Agreement Nos. HR00112290021 and HR001121C0168.
\bibliography{custom,references}

\appendix
\section{Reddit Data}
\subsection{Subreddits}
\label{app:subreddit}

We briefly introduce each subreddit in Table \ref{tab:subreddit}. 
Statistics of data from different subreddits are shown in Table \ref{tab:statistics}.

\begin{table*}[ht]
\small
\centering
\begin{tabular}{l|l}
\hline
\textbf{Subreddit}    & \textbf{Description}                                                                                        \\ \hline
\textbf{r/Liberal}        & \begin{tabular}[c]{@{}l@{}}r/Liberal is a subreddit dedicated to discussions and news from a liberal political \\perspective, focusing on progressive policies, social justice, and left-leaning viewpoints.\end{tabular}                                                        \\ \hline
\textbf{r/NeutralPolitics}    & \begin{tabular}[c]{@{}l@{}}r/NeutralPolitics is a community dedicated to evenhanded, empirical discussion of\\ political issues. It is a space to discuss policy and the tone of political debate.\end{tabular}                                                        \\ \hline
\textbf{r/Anarcho\_Capitalism} & \begin{tabular}[c]{@{}l@{}}r/Anarcho\_Capitalism is a subreddit dedicated to discussing \\ free-market capitalist anarchism and related topics, advocating for a society where \\ voluntary interactions enhance liberty and opportunity for all.\end{tabular} \\ \hline
\textbf{r/Conservative}        & \begin{tabular}[c]{@{}l@{}}r/Conservative offers the largest space on Reddit for fiscal and social \\ conservatives to explore and discuss political and cultural issues from a distinctly \\conservative perspective.\end{tabular}                         \\ \hline
\textbf{r/AskThe\_Donald}          & \begin{tabular}[c]{@{}l@{}}r/AskThe\_Donald is a subreddit that serves as a hub for supporters of former President \\Donald Trump, fostering discussions around conservative politics, pro-Trump views, \\ and right-wing ideologies.\end{tabular}                                                                                           \\ \hline
\hline
\textbf{r/keto}                & \begin{tabular}[c]{@{}l@{}}r/keto is a community for sharing experiences and advice about the low-carb Ketogenic \\diet, which supports a range of health conditions from diabetes to epilepsy\end{tabular}                                                  \\ \hline
\textbf{r/WeightLossAdvice}           & \begin{tabular}[c]{@{}l@{}}r/WeightLossAdvice is a subreddit where users share tips, strategies, and support for healthy \\ and sustainable weight loss, with a focus on practical advice and personal experiences.\end{tabular}                                                                                                        \\ \hline
\textbf{r/EDAnonymous}     & \begin{tabular}[c]{@{}l@{}}r/EDAnonymous is a subreddit that provides a supportive, anonymous space for individuals \\struggling with eating disorders to share their experiences, seek advice, and offer \\encouragement on the path to recovery.   \end{tabular}                            \\ \hline                        
\end{tabular}
\caption{Reddit forums used in this study.}
\label{tab:subreddit}
\end{table*}

\begin{table*}[ht]
\centering
\begin{tabular}{lcccll}
\hline
                           & \textbf{r/Lib}  & \textbf{r/NeutralPol} & \textbf{r/Anarcho\_Cap} & \multicolumn{1}{c}{\textbf{r/Conserv}} & \multicolumn{1}{c}{\textbf{r/ATDonald}} \\ \cline{2-6} 
\textbf{\# of comments}    & 31,233          & 35,725                & 670,686                 & \multicolumn{1}{c}{2,243,842}          & \multicolumn{1}{c}{142,543}             \\
\textbf{\# of submissions} & 5,243           & 2,072                 & 22,551                  & \multicolumn{1}{c}{153,813}            & \multicolumn{1}{c}{15,645}              \\ \cline{2-6} 
                           & \textbf{r/keto} & \textbf{r/WLAdvice}   & \textbf{r/EDAnonymous}  &                                        &                                         \\ \cline{1-4}
\textbf{\# of comments}    & 603,466         & 225,635               & 347,477                 &                                        &                                         \\
\textbf{\# of submissions} & 49,571          & 42,840                & 99,925                  &                                        &                                         \\ \hline
\end{tabular}
\caption{Number of comments and submissions in each subreddit.}
\label{tab:statistics}
\end{table*}

\section{Instructional Data Generation}
\subsection{Topic Modeling with BERTopic}
\label{app:bertopic}
Following the pipeline of BERTopic \cite{grootendorst2022bertopic}, we first obtain the text embeddings using \emph{all-mpnet-base-v2} \cite{song2020mpnet}. The embedding dimensions are reduced with UMAP, with n\_neighbors 15 and n\_components 5. Then the embeddings are clustered using HDBSCAN, with min\_cluster\_size 40. After fitting the model, we remove texts that are not assigned any topics. Table \ref{tab:topics-politics} and Table \ref{tab:topics-fitness} present the top 10 most frequent topics in \emph{politics} and \emph{diet}.

{\footnotesize
\begin{table*}[ht]
\addtolength{\tabcolsep}{-2.0pt}
\centering
\small
\begin{tabular}{cl}
\hline
\textbf{idx} & \multicolumn{1}{c}{\textbf{Topic Keywords}} \\ \hline
\textbf{1} & vaccine, covid, vaccinated, vaccines, flu, unvaccinated, vaccination, mrna, pandemic, data  \\
\textbf{2} & climate, prices, solar, change, co2, emissions, coal, cars, fuels, earth  \\
\textbf{3} & ballots, fraud, election, voter, ballot, evidence, voters, machines, audits, rigged  \\
\textbf{4} & abortion, abortions, fetus, roe, murder, unborn, conception, choice, womb, cells \\
\textbf{5} & ukraine, russia, putin, nato, war, ukrainians, crimea, invade, conflict, eu  \\
\textbf{6} & twitter, facebook, musk, google, social, platforms, companies, users, censorship, ban  \\
\textbf{7} & gun, guns, firearms, shootings, firearm, rifles, laws, amendment, ammo, armed  \\
\textbf{8} & impeachment, mueller, fbi, impeach, documents, collusion, comey, congress, whistleblower, crimes \\
\textbf{9} & gender, trans, transgender, lgbt, dysphoria, children, sexuality, identity, pronouns, genders\\
\textbf{10} & capitol, blm, antifa, riots, riot, insurrection, rioters, protesters, terrorism, january\\ \hline
\end{tabular}
\addtolength{\tabcolsep}{2.0pt}
\caption{Top-10 most frequent topics in \emph{politics}. Each topic is represented by the topic-10 keywords.}
\label{tab:topics-politics}
\end{table*}
}

\begin{table*}[ht]
\centering
\small
\begin{tabular}{cl}
\hline
\textbf{idx} & \multicolumn{1}{c}{\textbf{Topic Keywords}} \\ \hline
\textbf{1} & keto, protein, ketosis, started, macros, back, weeks, carbs, diet, calories  \\
\textbf{2} & skinny, thin, girls, like, overweight, body, underweight, weight, hate, myself\\
\textbf{3} & thank, therapy, relapse, recover, help, life, support, yourself, proud, treatment  \\
\textbf{4} & scale, week, water, lose, weighing, daily, trend, fluctuates, months, plateau \\
\textbf{5} & potassium, sodium, magnesium, electrolytes, mg, supplement, ketoade, citrate, 5000mg, chloride  \\
\textbf{6} & alcohol, drinking, beer, drink, drinks, liquor, alcoholic, gin, booze, drinker  \\
\textbf{7} & protein, macros, fat, min, kcal, 20g, carbs, bf, lean, need  \\
\textbf{8} & binging, restricting, binges, eating, hunger, control, bingeing, cycle, guilt, feeling \\
\textbf{9} & ed, eds, recovery, recover, therapist, treatment, me, help, coping, talk \\
\textbf{10} & coffee, tea, chocolate, brew, creamer, starbucks, latte, unsweetened, flavors, sweetener\\ \hline
\end{tabular}
\caption{Top-10 most frequent topics in \emph{diet}. Each topic is represented by the topic-10 keywords.}
\label{tab:topics-fitness}
\end{table*}

\subsection{Prompting Template for \textsc{CommInst}}
\label{app:prompt_temp_oe}

\begin{figure*}[ht]
    \centering
    \includegraphics[width=0.96\textwidth]{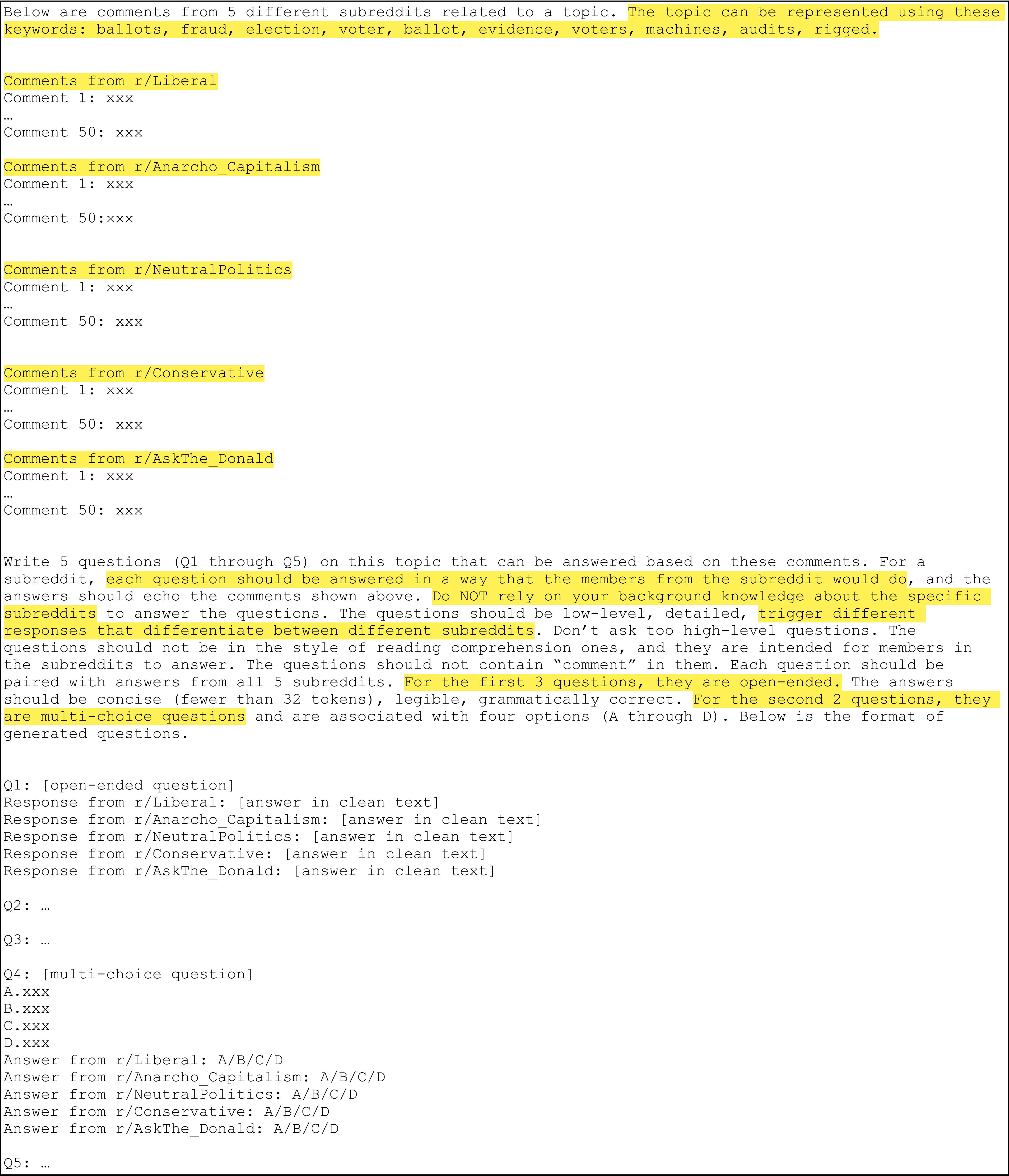}
    \caption{Prompting template to generate \textsc{CommInst} and \textsc{CommSurvey} in the \emph{politics} domain.}
    \label{fig:prompt_temp_oe}
\end{figure*}

The prompting template for generating the open-ended instruction-response pairs in \textsc{CommInst}  and multi-choice question-answer pairs in \textsc{CommSurvey} is shown in Figure \ref{fig:prompt_temp_oe}.

\subsection{Statistics of \textsc{CommInst} and \textsc{CommSurvey}}
\label{app:stats_comminst}

Table \ref{tab:statistics-inst} describes the number of open-ended instruction-response pairs in \textsc{CommInst} and multi-choice question-answer pairs in \textsc{CommSurvey}. The variance is due to their different topic coverage. 

\begin{table*}[ht]
\centering
\addtolength{\tabcolsep}{-3.0pt}
\begin{tabular}{lcccll}
\hline
                           & \textbf{r/Lib}  & \textbf{r/NeutralPol} & \textbf{r/Anarcho\_Cap} & \multicolumn{1}{c}{\textbf{r/Conserv}} & \multicolumn{1}{c}{\textbf{r/ATDonald}} \\ \cline{2-6} 
\textbf{\# of instructions}    & 234             & 234                   & 300                     & \multicolumn{1}{c}{303}                & \multicolumn{1}{c}{303}                 \\
\textbf{\# of survey questions} & 156             & 156                   & 200                     & \multicolumn{1}{c}{202}                & \multicolumn{1}{c}{202}                 \\ \cline{2-6} 
                           & \textbf{r/keto} & \textbf{r/WLAdvice}   & \textbf{r/EDAnonymous}  &                                        &                                         \\ \cline{1-4}
\textbf{\# of instructions}    & 1,032           & 885                   & 870                     &                                        &                                         \\
\textbf{\# of survey questions} & 688             & 590                   & 580                     &                                        &                                         \\ \hline
\end{tabular}
\addtolength{\tabcolsep}{-3.0pt}
\caption{Number of generated open-ended instruction-response pairs in \textsc{CommInst}$_i$ and multi-choice question-answer pairs in \textsc{CommSurvey}$_i$, for each community.}
\label{tab:statistics-inst}
\end{table*}

\subsection{Community Agreement Analysis}
\label{app:comm_agreement}
\begin{figure*}[th]
    \centering
    \begin{subfigure}[b]{0.49\textwidth}
        \centering
        \includegraphics[width=\textwidth]{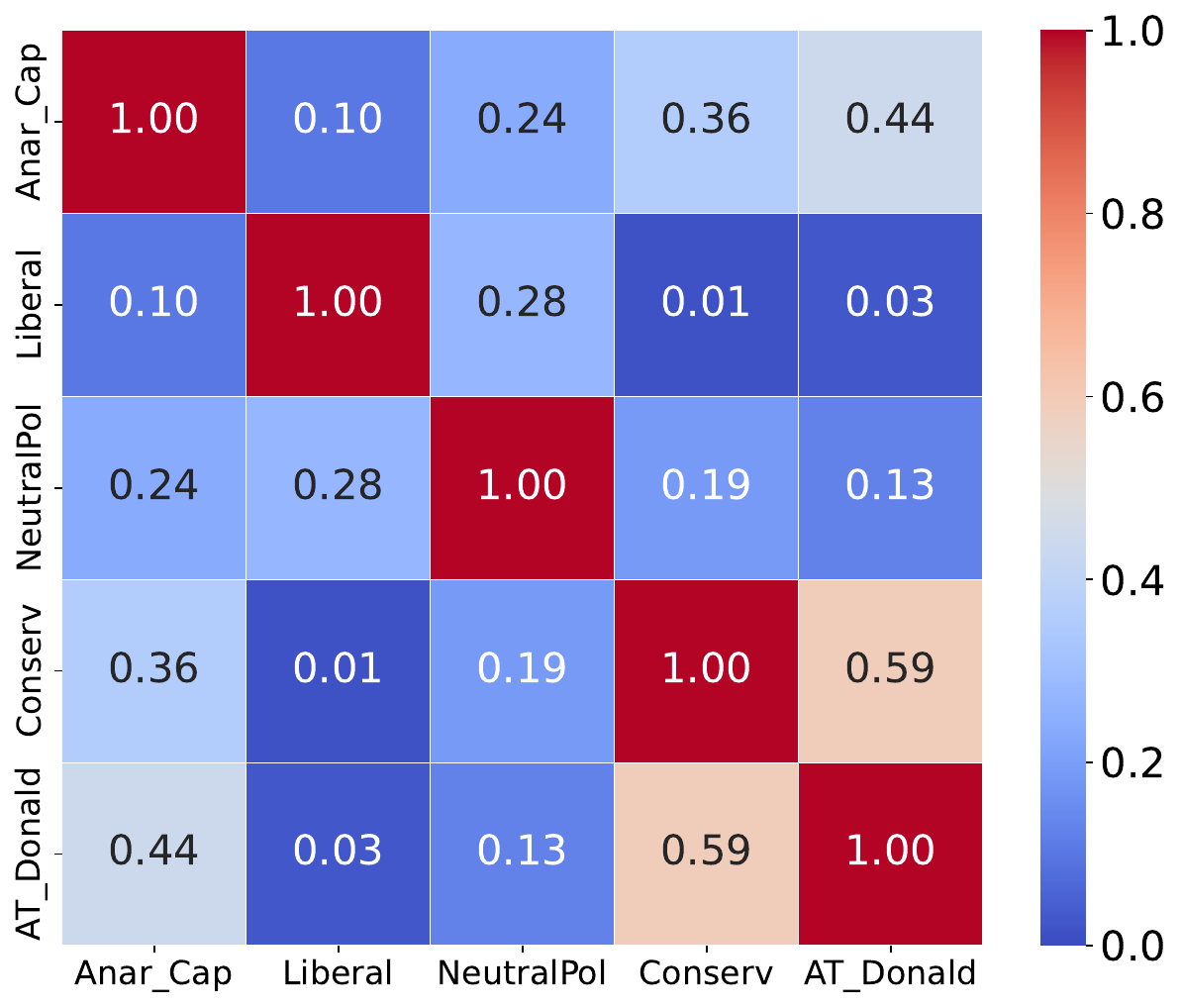}
        \caption{politics}
    \end{subfigure}
    \hfill
    \begin{subfigure}[b]{0.49\textwidth}
        \centering
        \includegraphics[width=\textwidth]{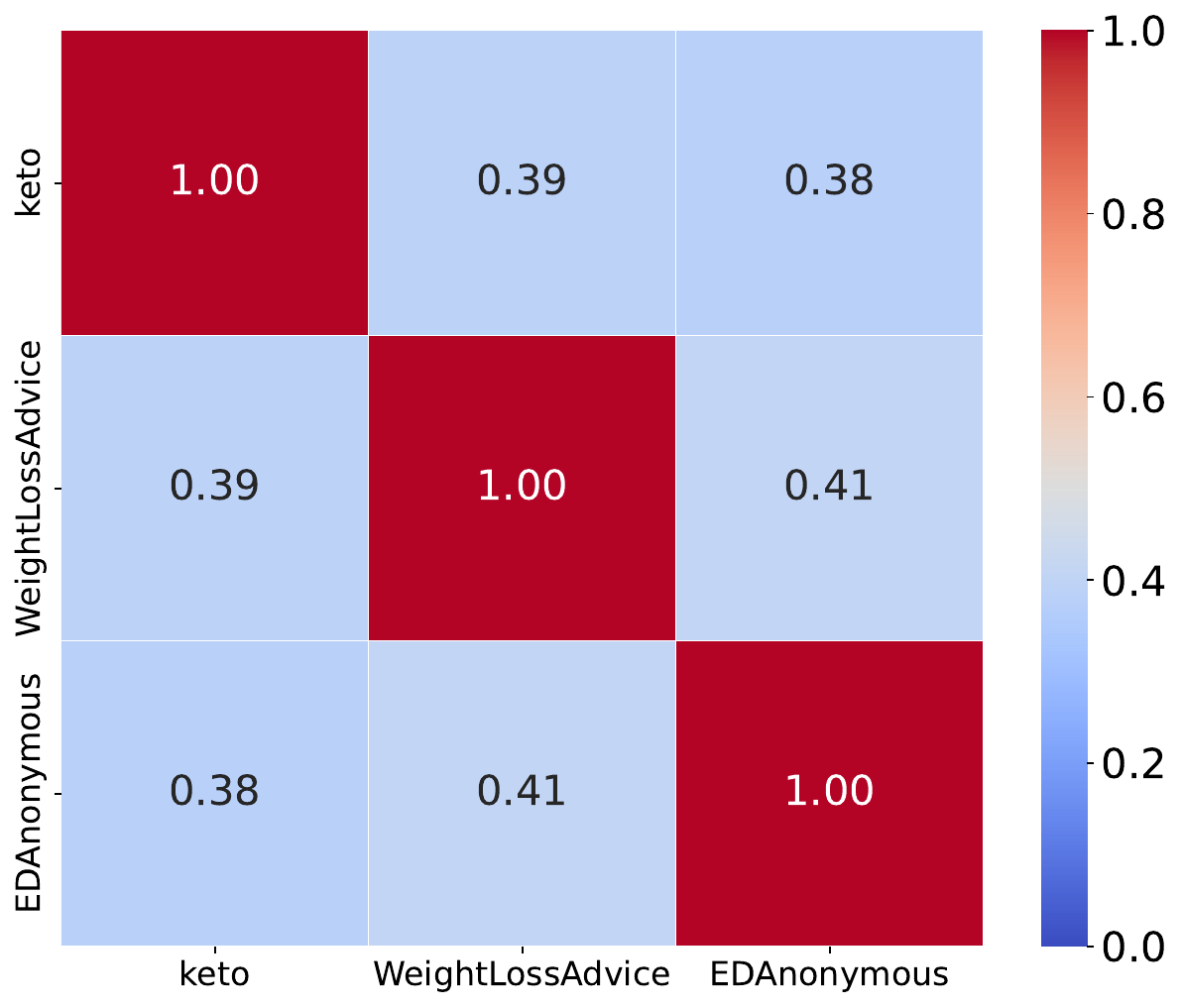}
        \caption{diet}
    \end{subfigure}
    \caption{Pairwise agreement between different communities, measured by Cohen's Kappa.}
    \label{fig:agreement_mat}
\end{figure*}

For a pair of communities $C_i$ and $C_j$, we identify their common survey questions in \textsc{CommSurvey}$_i$ and \textsc{CommSurvey}$_j$, and compute their agreement on the questions using Cohen's Kappa. The agreement matrices for \emph{politics} and \emph{diet} are shown in Figure \ref{fig:agreement_mat}.

In the \emph{politics} domain, we observe significant polarization between communities. For example, the agreement between \emph{r/Liberal} and both \emph{r/Conservative} (0.01) and \emph{r/AskThe\_Donald} (0.03) is extremely low, indicating the deep ideological divide between these left-leaning and right-leaning communities. This low agreement reflects how vastly different their responses are to the same political survey questions, aligning with the broader patterns of political polarization seen across online platforms. On the other hand, \emph{r/Conservative} and \emph{r/AskThe\_Donald} exhibit much higher agreement (0.59), highlighting the ideological overlap between these conservative-leaning communities, likely reflecting their shared values and viewpoints on key political issues.

Meanwhile, \emph{r/NeutralPolitics} shows moderate agreement with both \emph{r/Anarcho\_Capitalism} (0.24) and \emph{r/Liberal} (0.28), suggesting that as a centrist or neutral forum, it shares certain perspectives with both libertarian and liberal ideologies. Additionally, \emph{r/Anarcho\_Capitalism} shows higher agreement with conservative-leaning subreddits, such as \emph{r/Conservative} (0.36) and \emph{r/AskThe\_Donald} (0.44), reflecting shared economic or libertarian principles, despite divergences on other political issues. 

In the \emph{diet} domain, the relationships between the communities are more practical and less polarized compared to \emph{politics}. Communities like \emph{r/keto} and \emph{r/WeightLossAdvice} exhibit moderate agreement (0.39), likely reflecting shared goals related to weight loss and health, even though they may emphasize different strategies. Similarly, \emph{r/EDAnonymous} and \emph{r/keto} (0.38) show moderate overlap, suggesting that while their focus areas are different—one being centered on eating disorders and the other on ketogenic diets—they share some common ground in their approaches to health and diet-related topics.

The strongest alignment within the \emph{diet} domain is between \emph{r/WeightLossAdvice} and \emph{r/EDAnonymous} (0.41), which likely stems from overlapping concerns about diet and body image, which are central to both communities' discussions.

\section{Human Annotation}
\label{app:human_anno}
We verify the faithfulness of \textsc{CommSurvey} as the ``semi-ground truths'' for the communities via human annotation. We also tried directly posting the survey questions to the subreddits and having community members answer them. However, our posts were immediately removed by moderators and our accounts were banned from certain subreddits. This further indicates the difficulty of directly surveying online communities. 

\subsection{Annotator Recruiting}
The four annotators for the \emph{politics} domain closely follow political news and are knowledgeable about the platforms of the parties represented in our sample. The three annotators for the \emph{diet} domain are active members of diet and diet and have substantial knowledge of diet-related topics and trends.
To ensure the annotators were capable of accurately answering the survey questions for different communities, we implemented a thorough selection and training process. Annotators were chosen based on their demonstrated expertise and familiarity with the relevant domains. Additionally, we conducted preliminary tests where annotators answered a subset of questions, and their responses were evaluated for consistency and accuracy.

\subsection{Annotation Process}
\begin{figure*}
    \centering
    \includegraphics[width=0.9\textwidth]{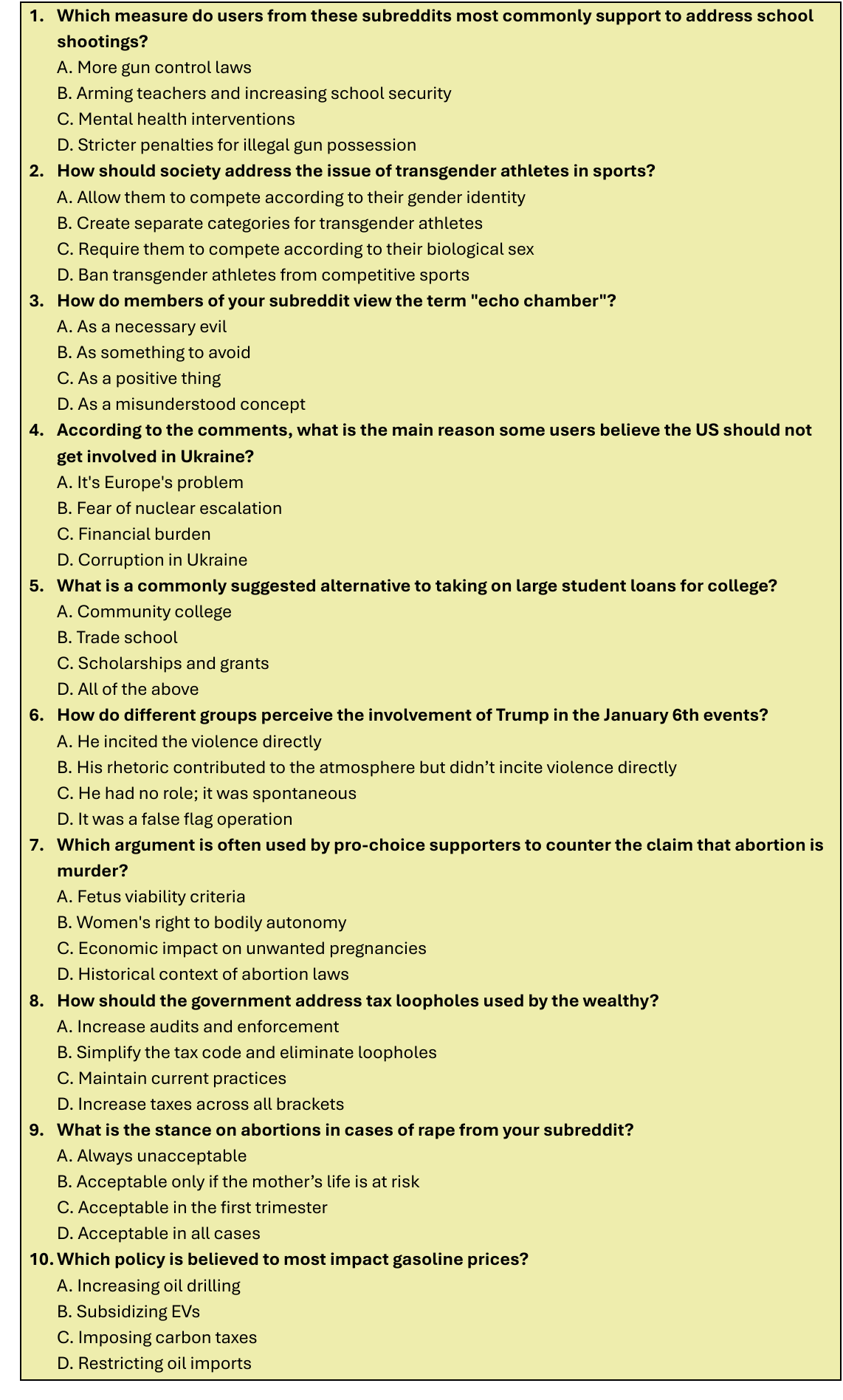}
    \caption{Ten multiple choice survey questions (out of the twenty) in \textsc{CommSurvey} for human evaluation in the \emph{politics} domain.}
    \label{fig:politics_survey}
\end{figure*}

\begin{figure*}
    \centering
    \includegraphics[width=0.95\textwidth]{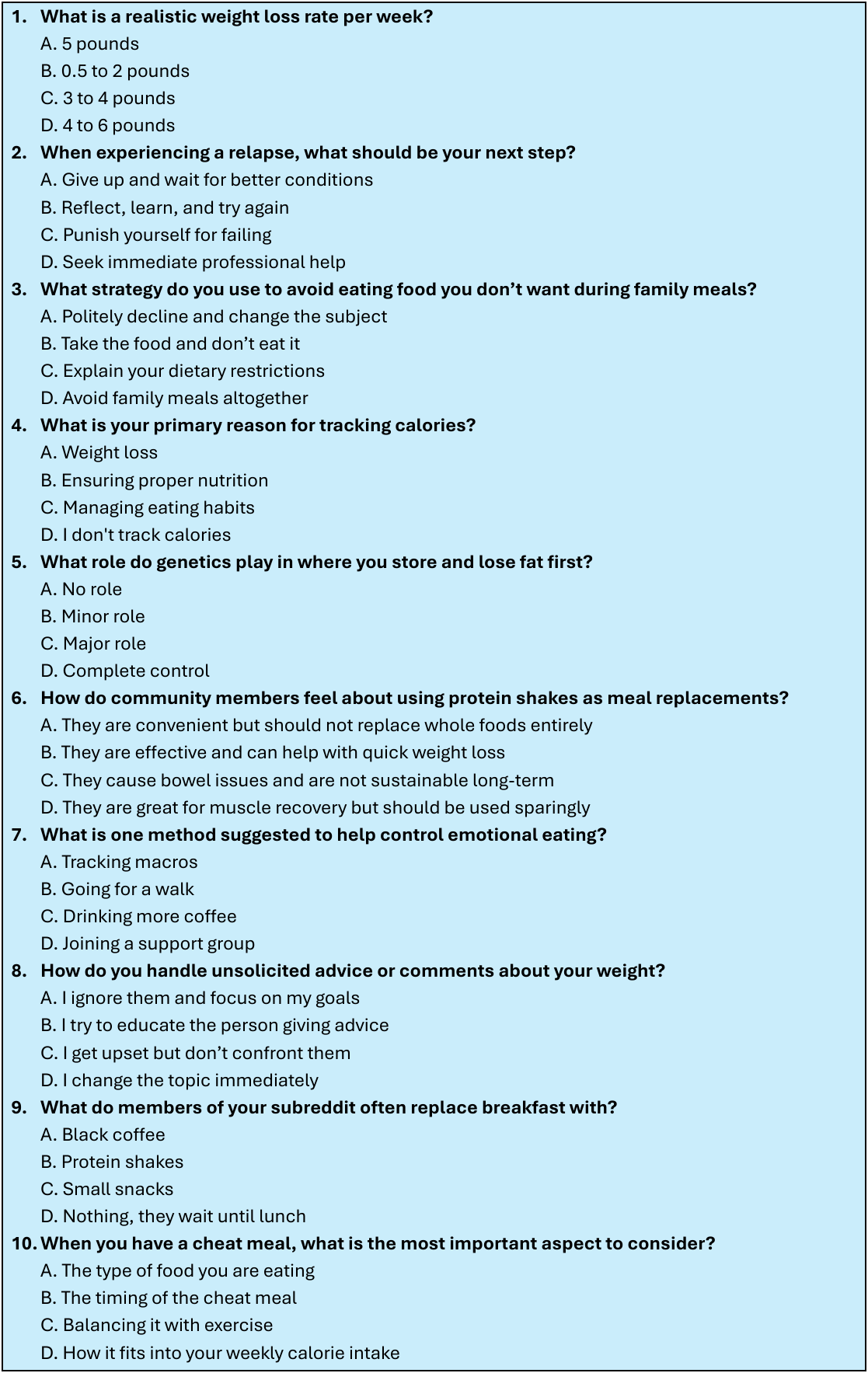}
    \caption{Ten multiple choice survey questions \textsc{CommSurvey} for human evaluation in the \emph{diet} domain.}
    \label{fig:fitness_survey}
\end{figure*}

We randomly sampled 20 questions from \textsc{CommSurvey-Politics} (Figure \ref{fig:politics_survey}) and 10 questions from \textsc{CommSurvey-Diet} (Figure~\ref{fig:fitness_survey}) for the annotators to answer. When answering each question, the annotators were instructed to search the relevant posts from the subreddit and learn their views from the discussions, instead of solely relying on their pre-assumption about the community.
Each annotator filled out the question for a different subset of the communities, as shown in Table \ref{tab:annotator_comm}. If multiple annotators annotated for a community, they had another round of discussions to resolve their disagreement. As a result, for community, the annotators delivered one set of annotations of the survey questions.

\subsection{Annotation Evaluation}
\label{sec:human_eval}
\begin{table*}[ht]
\centering
\addtolength{\tabcolsep}{-3.0pt}
\begin{tabular}{lcccll}
\hline
           & \multicolumn{1}{l}{\textbf{r/Lib}}  & \multicolumn{1}{l}{\textbf{r/NeutralPol}} & \multicolumn{1}{l}{\textbf{r/Anarcho\_Cap}} & \textbf{r/Conserv}             & \textbf{r/ATDonald}            \\ \cline{2-6} 
\textbf{Annotator 1} &                            &                                  & x                                  & \multicolumn{1}{c}{x} & \multicolumn{1}{c}{}  \\
\textbf{Annotator 2} & x                          &                                  & x                                  & \multicolumn{1}{c}{x} & \multicolumn{1}{c}{x} \\
\textbf{Annotator 3} & x                          &                                  & x                                  & \multicolumn{1}{c}{}  & \multicolumn{1}{c}{}  \\
\textbf{Annotator 4} & x                          &                                  &                                    & \multicolumn{1}{c}{x} & \multicolumn{1}{c}{}  \\ \hline
           & \multicolumn{1}{l}{\textbf{r/keto}} & \multicolumn{1}{l}{\textbf{r/WLAdvice}}   & \multicolumn{1}{l}{\textbf{r/EDAnonymous}}  &                       &                       \\ \cline{2-4}
\textbf{Annotator 1} & x                          & x                                & x                                  &                       &                       \\
\textbf{Annotator 2} & x                          &                                  & x                                  &                       &                       \\
\textbf{Annotator 3} &                            & x                                & x                                  &                       &                       \\ \hline
\end{tabular}
\addtolength{\tabcolsep}{-3.0pt}
\caption{Communities that each annotator annotated for.}
\label{tab:annotator_comm}
\end{table*}

For each subreddit, we compare the LLM-generated ``semi-ground truths'' to the label from human annotators, and present the accuracy in Table \ref{tab:human-eval}. In both domains, ``semi-ground truths'' achieve strong agreement with human annotations for most subreddits, which gives confidence that the advanced LLM-generated survey answers can be used as ``semi-ground truth'' to evaluate finetuned foundational LLMs.

\begin{table*}[ht]
\centering
\addtolength{\tabcolsep}{-3.0pt}
\begin{tabular}{llllll}
\hline
                                  & \textbf{r/Lib}                    & \textbf{r/NeutralPol}            & \textbf{r/Anarcho\_Cap}           & \textbf{r/Conserv}                & \textbf{r/ATDonald}               \\ \cline{2-6} 
\multicolumn{1}{c}{\textbf{Human-LLM Acc}} & \multicolumn{1}{c}{0.75} & \multicolumn{1}{c}{NA}  & \multicolumn{1}{c}{0.65} & \multicolumn{1}{c}{0.55} & \multicolumn{1}{c}{0.55} \\
                                  & \textbf{r/keto}                   & \textbf{r/WLAdvice}              & \textbf{r/EDAnonymous}            &                          &                          \\ \cline{2-4}
\multicolumn{1}{c}{\textbf{Human-LLM Acc}} & \multicolumn{1}{c}{0.7}  & \multicolumn{1}{c}{0.6} & \multicolumn{1}{c}{0.5}  &                          &                          \\ \hline
\end{tabular}
\addtolength{\tabcolsep}{3.0pt}
\caption{Agreement between human annotators' survey responses and the advanced LLM (GPT-4o) generated survey answers, measured by accuracy.}
\label{tab:human-eval}
\end{table*}

\end{document}